%% file: Main.tex
\documentclass[sigconf]{acmart}

\AtBeginDocument{%
  }

\copyrightyear{2025}
\acmYear{2025}
\setcopyright{acmlicensed}\acmConference[CIKM '25]{Proceedings of the 34th ACM International Conference on Information and Knowledge Management}{November 10--14, 2025}{Seoul, Republic of Korea}
\acmBooktitle{Proceedings of the 34th ACM International Conference on Information and Knowledge Management (CIKM '25), November 10--14, 2025, Seoul, Republic of Korea}
\acmDOI{10.1145/3746252.3761282}
\acmISBN{979-8-4007-2040-6/2025/11}
\acmSubmissionID{cfp6865}

\input{chapters/Preamble}

\begin{document}

\title{Leveraging Vulnerabilities in Temporal Graph Neural Networks via Strategic High-Impact Assaults}

\author{Dong Hyun Jeon}
\orcid{0009-0000-9046-0470}
\affiliation{%
  \institution{Bowling Green State University}
  \streetaddress{Hayes 227}
  \city{Bowling Green}
  \state{Ohio}
  \country{USA}
}
\email{djeon@bgsu.edu}

\author{Lijing Zhu}
\orcid{0009-0002-7107-2880}
\affiliation{%
  \institution{University of Houston-Clear Lake}
  \city{Houston}
  \state{Texas}
  \country{USA}
  }
\email{zhul@uhcl.edu}

\author{Haifang	Li}
\orcid{0009-0003-1098-4023}
\affiliation{%
  \institution{Mayo Clinic}
  \city{Jacksonville}
  \state{Florida}
  \country{USA}
  }
\email{li.haifang@mayo.edu}

\author{Pengze Li}
\orcid{0000-0001-7015-0491}
\affiliation{%
  \institution{Mayo Clinic}
  \city{Jacksonville}
  \state{Florida}
  \country{USA}
  }
\email{li.pengze@mayo.edu}

\author{Jingna Feng}
\orcid{0000-0002-0434-2513}
\affiliation{%
  \institution{Mayo Clinic}
  \city{Jacksonville}
  \state{Florida}
  \country{USA}
  }
\email{feng.jingna@mayo.edu}

\author{Tiehang	Duan}
\orcid{0000-0003-4323-642X}
\affiliation{%
  \institution{Grand Valley State University}
  \city{Allendale}
  \state{Michigan}
  \country{USA}
  }
\email{tiehang.duan@gmail.com}

\author{Houbing Herbert	Song}
\orcid{0000-0003-2631-9223}
\affiliation{%
  \institution{University of Maryland at Baltimore County}
  \city{Baltimore}
  \state{Maryland}
  \country{USA}
  }
\email{songh@umbc.edu}

\author{Cui Tao}
\orcid{0000-0002-4267-1924}
\affiliation{%
  \institution{Mayo Clinic}
  \city{Jacksonville}
  \state{Florida}
  \country{USA}
  }
\email{tao.cui@mayo.edu}

\author{Shuteng	Niu}
\orcid{0000-0002-1069-9236}
\affiliation{%
  \institution{Mayo Clinic}
  \city{Jacksonville}
  \state{Florida}
  \country{USA}
  }
\email{niu.shuteng@mayo.edu}


\renewcommand{\shortauthors}{Dong Hyun Jeon et al.}

\input{chapters/Abstract}



\maketitle

\input{chapters/Introduction}
\input{chapters/RelatedWork}
\input{chapters/ProblemFormulation}

\input{chapters/Methodology}
\input{chapters/Experiments}
\input{chapters/Ablation}
\input{chapters/Conclusion}
\input{chapters/GenAIDisclosure}

\bibliographystyle{ACM-Reference-Format}
\balance
\bibliography{Reference}

\end{document}

%% file: chapters/Preamble.tex
\usepackage{amsmath,amsfonts}
\usepackage{algorithmic}
\usepackage{bm}
\usepackage{multirow}
\usepackage[ruled, lined, linesnumbered, commentsnumbered, longend]{algorithm2e}

\def\secref#1{Section~\ref{#1}}
\def\figref#1{Fig.~\ref{#1}}
\def\tabref#1{Table~\ref{#1}}
\def\eqref#1{Eq.~\ref{#1}}


%% file: chapters/Abstract.tex
\begin{abstract}

Temporal Graph Neural Networks (TGNNs) have become indispensable for analyzing dynamic graphs in critical applications such as social networks, communication systems, and financial networks. However, the robustness of TGNNs against adversarial attacks, particularly sophisticated attacks that exploit the temporal dimension, remains a significant challenge. Existing attack methods for Spatio-Temporal Dynamic Graphs (STDGs) often rely on simplistic, easily detectable perturbations (e.g., random edge additions/deletions) and fail to strategically target the most influential nodes and edges for maximum impact. We introduce the High Impact Attack (HIA), a novel restricted black-box attack framework specifically designed to overcome these limitations and expose critical vulnerabilities in TGNNs. HIA leverages a data-driven surrogate model to identify structurally important nodes (central to network connectivity) and dynamically important nodes (critical for the graph's temporal evolution). It then employs a hybrid perturbation strategy, combining strategic edge injection (to create misleading connections) and targeted edge deletion (to disrupt essential pathways), maximizing TGNN performance degradation. Importantly, HIA minimizes the number of perturbations to enhance stealth, making it more challenging to detect. Comprehensive experiments on five real-world datasets and four representative TGNN architectures (TGN, JODIE, DySAT, and TGAT) demonstrate that HIA significantly reduces TGNN accuracy on the link prediction task, achieving up to a 35.55\% decrease in Mean Reciprocal Rank (MRR) – a substantial improvement over state-of-the-art baselines. These results highlight fundamental vulnerabilities in current STDG models and underscore the urgent need for robust defenses that account for both structural and temporal dynamics. Code and Data are available at \url{https://github.com/ryandhjeon/hia}.

\end{abstract}

%% file: chapters/Introduction.tex
\section{Introduction}

The increasing prevalence of dynamic graphs in critical applications, from social networks~\cite{fan2019graph, hussain2021structack} and recommendation systems~\cite{jeon2024kgif, fan2019graph} to fraud detection~\cite{pourhabibi2020fraud, sun2022reinforced, zeager2017adversarial, fursov2021adversarial} and cybersecurity~\cite{bowman2021towards, liu2022recent, pingle2019relext}, demands robust and reliable analysis tools. Spatio-Temporal Graph Neural Networks (STGNNs), which model graphs where both node features and structure evolve, have emerged as powerful methods for capturing the complex, time-evolving relationships within these dynamic systems~\cite{zhao2024adversarial, sharma2023temporal, zhu2024flexible, pareja2020evolvegcn}. However, like other deep learning models, STGNNs are vulnerable to adversarial attacks: subtle, carefully crafted manipulations of the graph data that can drastically degrade performance~\cite{zugner2020adversarial, wu2019adversarial, jin2021adversarial}. Such vulnerabilities pose significant real-world threats. Understanding these through rigorous attack research is paramount for developing trustworthy Temporal Graph Neural Networks (TGNNs).

While adversarial attacks on static graphs are well-studied~\cite{zhu2019robust, zhang2020gnnguard}, the temporal dimension introduces unique challenges and opportunities. Unlike static attacks that manipulate a fixed structure, dynamic graph attacks must account for evolving node importance, temporal dependencies, and the need for stealth in a constantly changing environment~\cite{chen2021time, liu2023adversarial}. These complexities necessitate more sophisticated attack strategies that consider:

\begin{itemize}
    \item \textbf{Dynamic Vulnerabilities:} Node influence evolves, and perturbations must consider not just immediate impact but also how they influence future graph evolution and TGNN predictions, as minor changes at one time step can have cascading effects. Attacks thus need to adapt by targeting persistently or increasingly important nodes and accounting for these temporal dependencies.
    \item \textbf{Stealth in Dynamic Contexts:} While anomaly detection can be more challenging in dynamic graphs, the subtlety of perturbations is crucial for evading detection by blending with natural graph evolution.
    \item \textbf{Computational Scalability:} The inherent dynamics add processing overhead, necessitating computationally efficient attack generation, especially for large, evolving graphs.
\end{itemize}

\begin{figure}[t]
\centering
\includegraphics[width=0.8\columnwidth]{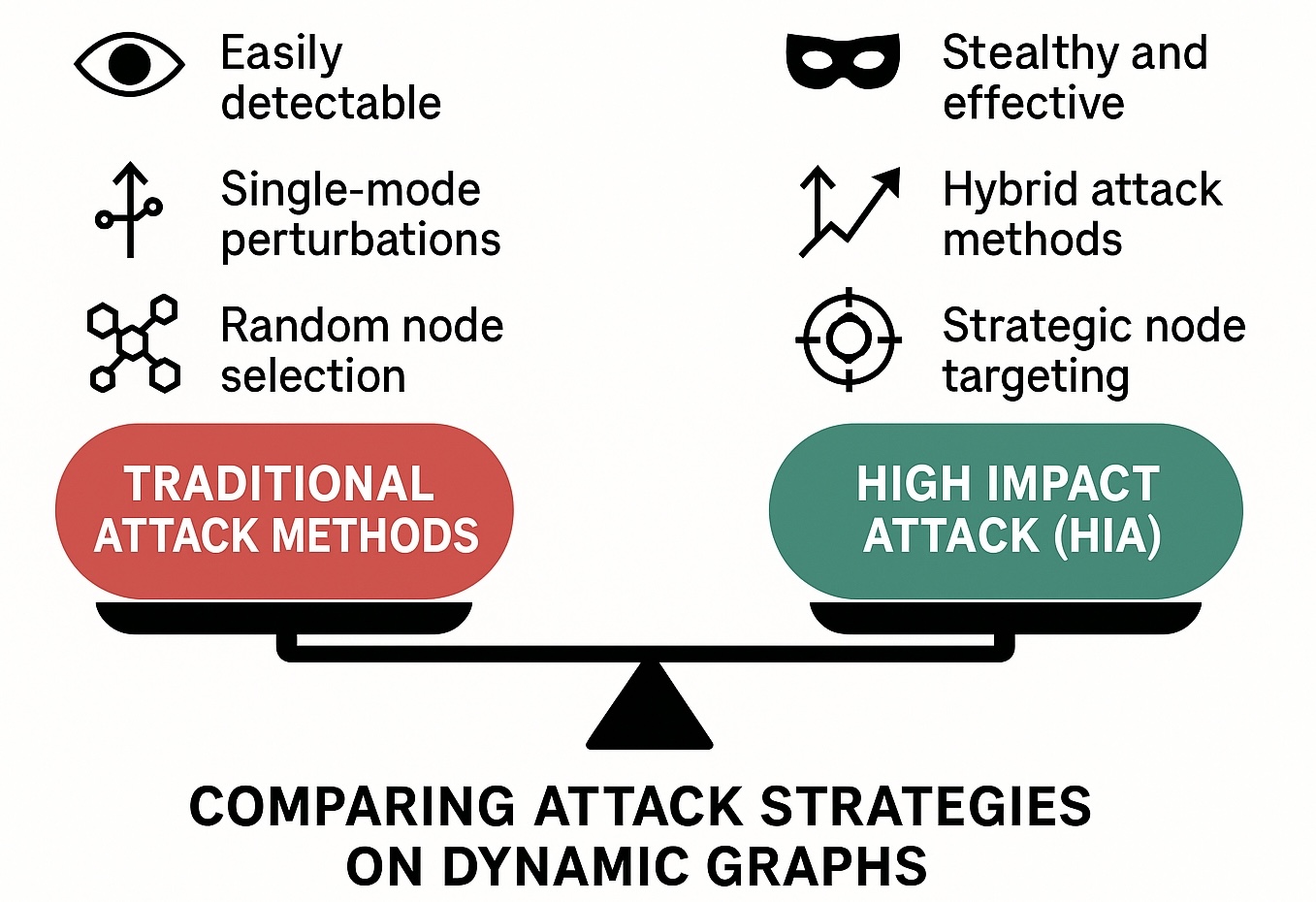}
\caption{Comparison between HIA and traditional methods, highlighting HIA's hybrid approach, strategic node targeting, and improved stealth. Traditional methods are often easily detectable and rely on single-mode perturbations and random node selection.}
\label{fig:hia-compare}
\end{figure}

Despite STGNNs' power, existing adversarial attacks often prove inadequate for temporal graphs. Many treat them as static snapshots or neglect the interplay of spatial and temporal factors, limiting their efficacy~\cite{zhao2024adversarial, lee2024ssaad}. Common methods use simplistic, detectable perturbations (e.g., random modifications~\cite{fursov2021adversarial, ma2020towards, zou2021tdgia}) that fail to capture the intricate, evolving dependencies TGNNs learn (\figref{fig:hia-compare}). This paper introduces strategies that address these shortcomings by explicitly considering temporal dynamics. Rigorous attack development is crucial: understanding TGNN vulnerabilities is a prerequisite for designing robust defenses.

We introduce the \textbf{H}igh \textbf{I}mpact \textbf{A}ttack (HIA), a novel \textbf{restricted black-box attack} to significantly improve existing attacks on STGNNs for link prediction in a poisoning setting. Without knowledge of victim TGNN internals (beyond task and some unperturbed training data), HIA's attacker strategically perturbs high-impact nodes—critical to predictions and temporal evolution—by adding/deleting few edges (budget \(\Delta\)). HIA employs a data-driven surrogate to guide a hybrid perturbation strategy (strategic injection/deletion), maximizing performance degradation while minimizing perturbations to enhance stealth. This approach leverages spatio-temporal dependencies to disrupt local and global graph dynamics, offering crucial insights for designing more resilient TGNNs.

To advance adversarial attack research on temporal graphs and inform future defenses, we make the following contributions:

\begin{enumerate}
    \item \textbf{Novel Dynamic Importance-Guided Attack Strategy:} HIA introduces a unique $\text{Impact}(v)$ score that identifies high-impact temporal graph nodes by synergistically integrating their \textit{temporal dynamics} (e.g., degree growth, \secref{sec:high_attraction_nodes}), \textit{structural centrality} (betweenness, \secref{sec:bridge_nodes}), and \textit{community context} (\secref{sec:community_aware_targeting}). This multi-faceted, evolving significance targeting directly addresses dynamic vulnerabilities, empirically outperforming temporally naive attacks (\secref{sec:Experiments}).
    \item \textbf{Scalable Limited-Knowledge Framework for Temporal Graphs:} HIA offers a practical framework for large, evolving graphs without internal victim model access, tackling computational scalability via: (i) a data-driven surrogate that obviates direct, costly victim model interactions for guidance; (ii) $\text{Impact}(v)$-guided selective targeting of minimal critical node subsets; and (iii) efficient graph metric computations. Its scalability and efficient attack generation are empirically validated.
    \item \textbf{State-of-the-Art Empirical Effectiveness and Robustness:} HIA's efficacy is demonstrated across five diverse datasets and four TGNN architectures, achieving superior attack performance with minimal perturbation budgets. This validates HIA's practical utility and its capacity to achieve dynamic stealth through impactful yet efficient perturbations.
\end{enumerate}

The paper is structured as follows: \secref{sec:RelatedWork} reviews related work; \secref{sec:ProblemFormulation} formally defines the problem and attack objectives; \secref{sec:Methodology} details the HIA framework; \secref{sec:Experiments} describes the experimental setup and evaluates HIA's performance; \secref{sec:ablation} justifies the results; and \secref{sec:Conclusion} summarizes findings and discusses future research.

%% file: chapters/RelatedWork.tex
\section{Related Work} \label{sec:RelatedWork}

\subsection{Targeting Influential Nodes}

Many studies have delved into adversarial attacks on static graphs~\cite{wu2020comprehensive, sun2022adversarial}. Pioneering work such as Nettack~\cite{zugner2018adversarial} introduced \textit{targeted poisoning attacks}, distinguishing between direct and influencer attacks. It employs a greedy algorithm with a linearized surrogate model to handle the discrete graph domain and proposes unnoticeable perturbations by preserving properties like degree distribution~\cite{zugner2018adversarial}. Subsequent gradient-based methods like FGA~\cite{chen2018fast} and PGD-based attacks~\cite{xu2019topology} offer more refined perturbation selection but face scalability issues. For global performance degradation, Metattack~\cite{zugner2020adversarial} formulates the attack as a bilevel optimization problem solved via meta-learning.

Scalability was addressed by SGA~\cite{li2021adversarial}, which improves efficiency by attacking a localized k-hop subgraph with an SGC surrogate. Alternative paradigms include RL-S2V~\cite{dai2018adversarial}, which uses reinforcement learning for sequential edge perturbations but faces high training complexity. The general surrogate-based approach has also proven effective for generating transferable attacks on heterogeneous graphs~\cite{zhao2024hgattack}.

A critical limitation unites these methods: they are designed for static graphs and rely on static importance measures. Static metrics like betweenness centrality~\cite{riondato2018abra} fail in dynamic contexts where influence evolves, as attacks on graph snapshots ignore temporal causality and can result in ineffective or detectable perturbations~\cite{lerman2010centrality, donnat2018tracking}. 

\subsection{Adversarial Attacks on Temporal Graphs}
Attacking temporal graphs introduces unique constraints, most notably the necessity of respecting \textit{time-respecting paths}, where information flows along edge sequences with non-decreasing timestamps~\cite{lerman2010centrality, tsalouchidou2020temporal, saxena2020centrality}. This temporal ordering requires that perturbations not only disrupt spatial graph structures but also remain causally consistent and temporally coherent to avoid easy detection.

Early attacks on temporal graphs treated them as sequences of static snapshots. \textbf{TGA}~\cite{chen2021time}, a white-box evasion attack, greedily selects perturbations on a per-timestamp basis but ignores cross-time dependencies, risking temporally incoherent modifications. To improve consistency, \textbf{TD-PGD}~\cite{sharma2023temporal} extends Projected Gradient Descent (PGD) by accumulating gradients across snapshots. However, both methods are white-box and require full model access, limiting their practical applicability.

More recent methods address continuous-time graphs and specialized scenarios. \textbf{T-SPEAR}~\cite{lee2024ssaad} is a stealthy poisoning attack for link prediction using distributed edge injections, but its reliance on white-box pre-computation limits its impact under tight budgets. \textbf{Adversparse}~\cite{li2022adversparse} targets forecasting tasks using ADMM for sparse perturbations, but its focus on minimizing edit norms over strategic placement limits its efficacy. Specialized attacks, such as those for traffic forecasting~\cite{sun2022adversarial}, are often task-specific and lack general scalability. Reinforcement learning-based approaches have also been explored but often generate erratic, detectable perturbations that lack temporal smoothness~\cite{nie2023reinforcement}.

%% file: chapters/ProblemFormulation.tex
\section{Problem Formulation}
\label{sec:ProblemFormulation}

We study limited-knowledge poisoning adversarial attacks on TGNNs, focusing on the link prediction task. The adversary aims to degrade model performance by adding ($\mathcal{E}_A$) and/or deleting ($\mathcal{E}_D$) edges before model training. Node modifications are not considered. The adversary's objective is formally defined as:

\begin{equation}
\label{eq:attacker_objective}
\begin{aligned}
    &\text{Maximize } \mathcal{L}(\mathcal{G}', \theta) \\
    &\text{Subject to: } |\mathcal{E}_A| + |\mathcal{E}_D| \leq \Delta,
\end{aligned}
\end{equation}
where $\mathcal{L}$ is the target model's loss function (unknown to the attacker), $\mathcal{G}'$ is the perturbed graph, $\theta$ represents the target's unknown parameters, and $\Delta$ is the perturbation budget, limiting the total number of added and deleted edges.

\paragraph{Attacker Model.}
HIA operates under a \textbf{restricted black-box attack} setting. The attacker has access to the historical unperturbed training graph data $\mathcal{G} = (\mathcal{V}, \mathcal{E})$ (including features and timestamps), which is used to train a local surrogate model. However, the attacker has no knowledge of the target victim model's architecture, internal parameters, or access to its predictions or true labels during the attack generation process. The attacker is aware that the task is link prediction and is limited to modifying only edges within a defined perturbation budget $\Delta$.

\paragraph{Discussion of Assumptions:} Assuming complete historical knowledge provides an upper bound on attack effectiveness, as real-world attackers might have only partial information.  Future work will investigate scenarios with limited historical data.  The perturbation budget, $\Delta$, is currently a fraction of the original graph's edges, implying uniform manipulation costs.  In practice, costs might vary (e.g., creating a new connection could be more detectable than deleting a weak one). Future work could explore heterogeneous edge manipulation costs.

\subsection{Graph Attack Overview}

Real-world graphs exhibit dynamic behavior, with edges appearing and disappearing over time. We employ a Sparse Event-Based Representation to capture this temporal evolution:

\begin{equation}
\label{eq:event_representation}
    \mathcal{E} = \{(u_1, v_1, t_1), (u_2, v_2, t_2), \dots, (u_m, v_m, t_m)\},
\end{equation}
where each triplet $(u, v, t)$ denotes an interaction between nodes $u$ and $v$ at time $t$ ($t_1 < t_2 < ... < t_m$). Duplicate edges are not permitted. This representation facilitates precise, time-aware adversarial modifications. The attack modifies the temporal edge list to introduce inconsistencies:

\begin{equation}
\label{eq:perturbed_edge_set_overview}
    \hat{\mathcal{E}} = (\mathcal{E} \setminus \mathcal{E}_D) \cup \mathcal{E}_A,
\end{equation}
where $\hat{\mathcal{E}}$ represents the perturbed edge set. The attack strategy combines Edge Injection (introducing misleading relationships) and Edge Deletion (eliminating crucial connections).

\subsection{Dynamic Graph Adversarial Attack}

A STDG is formally defined as:

\begin{equation}
\label{eq:stdg_definition}
    \mathcal{G} = (\mathcal{V}, \mathcal{E}),
\end{equation}
with $\mathcal{V}$ representing the set of nodes and $\mathcal{E}$ the set of temporal edges. TGNNs compute time-aware node embeddings, $h_u(t)$ and $h_v(t)$, to predict the likelihood of an interaction:

\begin{equation}
\label{eq:edge_likelihood}
    \hat{y}_{uvt} = \sigma\Big(\text{clf}\big(h_u(t), h_v(t)\big)\Big),
\end{equation}
where $\text{clf}(\cdot)$ is a function that scores edge formation likelihood, and $\sigma(\cdot)$ is the sigmoid function. TGNNs are typically trained by minimizing a loss function, such as binary cross-entropy:

\begin{equation}
\label{eq:loss_function}
\mathcal{L_E} =  - \sum_{(u,v,t) \in \mathcal{E}} \log(\hat{y}_{uvt}) - \sum_{n \in \mathcal{N}_{u,t}} \log(1-\hat{y}_{unt}).
\end{equation}
In this function, \(y_{uvt}\) represents the ground truth labels (1 for an existing edge, 0 otherwise), and \(\mathcal{N}_{u,t}\) is the set of negative samples for node \(u\) at time \(t\).

\begin{figure*}[htb]
    \includegraphics[width=\textwidth]{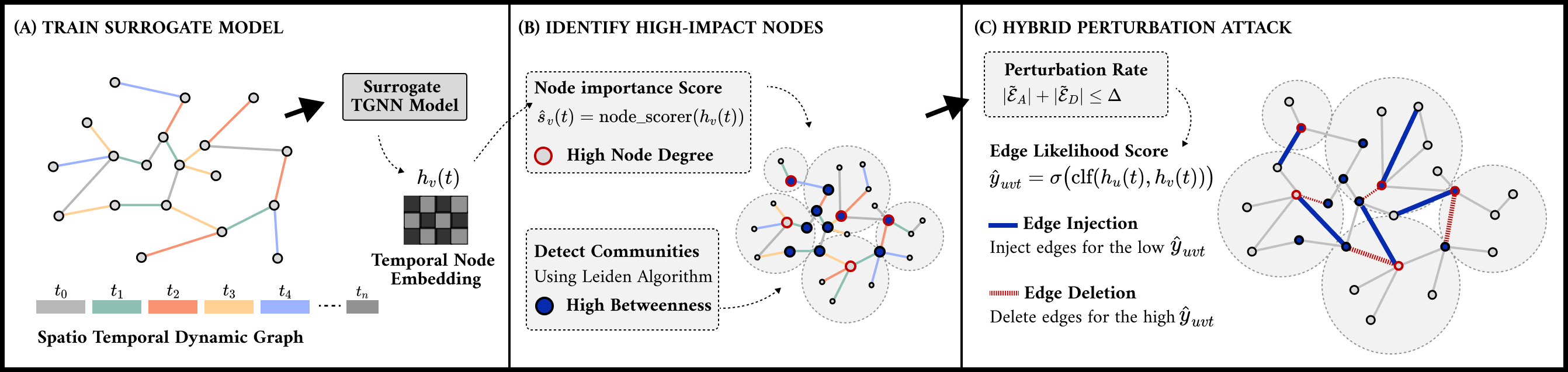}
    \caption{Overview of the HIA Strategy. The diagram illustrates the three core components: (A) Train Surrogate Model: A spatio-temporal dynamic graph is input to a surrogate TGNN, which learns temporal node embeddings. (B) Identify High-Impact Nodes: The surrogate model computes node importance scores and identifies communities. High-degree (red circles) and high-betweenness (blue circles) nodes within communities are targeted. (C) Hybrid Perturbation Attack: Using a predefined perturbation rate, the attack performs edge injection and deletion based on edge likelihood scores from the surrogate. Low-likelihood edges are injected (thick blue lines), and high-likelihood edges, especially those connected to high-impact nodes, are deleted (thick dashed red lines). The result is the perturbed graph.
}
    \label{fig:HIA}
\end{figure*}

%% file: chapters/Methodology.tex
\section{Proposed Model: HIA}
\label{sec:Methodology}

HIA is an adversarial framework designed to degrade TGNN performance by strategically perturbing high-impact nodes and edges—those most critical for the graph's structure and temporal evolution. This approach addresses the limitations of traditional, often random or uniform, and less effective or detectable perturbations. As depicted in \figref{fig:HIA}, HIA comprises three interconnected components:

\begin{enumerate}
    \item \textbf{Surrogate Model} (\secref{sec:surrogate}) approximates the target TGNN and is used to estimate node importance and edge likelihood.
    \item \textbf{Strategic Node Selection} (\secref{sec:node_selection}) identifies high-impact nodes based on the outputs of the surrogate model and the graph's community structure.
    \item \textbf{Hybrid Perturbation Strategy} (\secref{sec:perturbation}) combines targeted edge deletion and injection, guided by the surrogate model, to maximize the attack's impact while maintaining stealth.
\end{enumerate}

\subsection{Surrogate Model}
\label{sec:surrogate}

Due to restricted black-box access, we employ a surrogate model. Specifically, we adopt a Temporal Graph Network (TGN) architecture for this surrogate~\cite{lee2024ssaad}. Though potentially simpler in configuration (e.g., fewer layers or dimensions) than the victim, our surrogate TGN is trained on observable graph interactions to approximate victim behavior.

As depicted in \figref{fig:HIA}(A), this TGN surrogate learns lightweight temporal node embeddings $h_v(t)$ using its inherent memory (e.g., GRU-based) and temporal graph attention mechanisms to capture historical states and aggregate neighborhood information. These embeddings are then used to derive two crucial transferable priors. Edge likelihoods ($\hat{y}_{uvt}$), indicating link probabilities, are estimated by a classifier (e.g., an MLP or dot product with sigmoid) on the embeddings $h_u(t)$ and $h_v(t)$ from the surrogate TGN~\cite{lee2024ssaad}. Concurrently, node importance scores ($\hat{s}_v(t)$) are computed by applying a learned function (e.g., a small MLP) to individual embeddings $h_v(t)$; these scores quantify node impact and are used directly in HIA's perturbation strategy. 

This dual-output approach is critical. While graph-based metrics help identify structurally and dynamically important nodes, the surrogate model is essential for providing the nuanced, learned priors needed to execute the attack effectively. Specifically, the edge likelihood scores ($\hat{y}_{uvt}$) are indispensable for the hybrid perturbation strategy: they allow HIA to identify which existing edges the victim model likely considers important (and should thus be deleted) and which non-existent edges would be most disruptive to inject (i.e., those the model considers highly improbable).

\subsection{Strategic Node Selection}
\label{sec:node_selection}

HIA aims to maximize disruption while minimizing detectability by focusing on a small but critical set of nodes. As shown in \figref{fig:HIA}(B), to identify these nodes, HIA computes a holistic $\text{Impact}(v)$ score that uniquely integrates three dimensions of a node's importance: its dynamic influence captured by temporal growth, its role in global structure measured by centrality, and its local community context. The following subsections detail the components that constitute this score.

\subsubsection{High-Attraction Nodes}
\label{sec:high_attraction_nodes}

These nodes demonstrate rapid growth in their connectivity over time, a strong indicator of increasing influence within the network. We quantify this influence using the temporal degree growth rate, $\Delta d_u(t)$. Using only the degree at a single time point would be insufficient, as it does not capture the dynamics of node importance. A high degree at a single time step could represent a node that was previously important but is no longer influential, or a node that has consistently maintained high connectivity. The growth rate, however, highlights nodes that are actively gaining influence. A simple average of past degrees would be susceptible to legacy effects, potentially overemphasizing nodes whose importance is waning.

The temporal degree growth rate is mathematically defined as:

\begin{equation}
\Delta d_u(t) = \frac{d_u(t) - d_u(t - \Delta t)}{\Delta t},
\label{eq:temp_degree_growth_short}
\end{equation}
where $d_u(t)$ represents the degree of node $u$ at time $t$, and $\Delta t$ is a predefined time window. This metric focuses on the recent change in connectivity, making it a more sensitive indicator of a node's current and potential future importance. This is particularly crucial in the context of adversarial attacks, where we aim to target nodes that are likely to be important for future predictions made by the TGNN. This approach aligns with established findings in dynamic graph analysis, which emphasize the importance of growth rate in identifying influential nodes~\cite{kumar2019predicting, liben2003link, cai2005mining}. This forward-looking metric is therefore a cornerstone of the HIA framework, ensuring our attack prioritizes nodes that are not just historically important, but are becoming crucial to the graph's future evolution.

\subsubsection{Bridge Nodes}
\label{sec:bridge_nodes}

These nodes are essential for maintaining global connectivity within the graph, acting as connectors between different regions or communities. We quantify the importance of bridge nodes using betweenness centrality:

\begin{equation}
\label{eq:betweenness_centrality}
C_B(v) = \sum_{s,t \in \mathcal{V}} \frac{\sigma_{st}(v)}{\sigma_{st}},
\end{equation}
where $\sigma_{st}$ represents the number of shortest paths between nodes $s$ and $t$, and $\sigma_{st}(v)$ represents the number of those shortest paths that pass through node $v$. A high $C_B(v)$ value indicates that a node serves as a critical conduit for information flow across the network~\cite{hussain2021structack}. Disrupting such nodes can significantly fragment the graph and hinder the TGNN's ability to learn global relationships.

\subsubsection{Community-Aware Targeting}
\label{sec:community_aware_targeting}

In addition to the individual node metrics discussed in \secref{sec:high_attraction_nodes} and \secref{sec:bridge_nodes}, HIA incorporates the graph's community structure to further refine the target selection process. We employ the Leiden algorithm~\cite{traag2019louvain} for community detection. This algorithm partitions the graph by maximizing modularity \(Q\), defined as:
\begin{equation}
Q = \frac{1}{2m} \sum_{i,j} \left[ A_{ij} - \frac{k_i k_j}{2m} \right] \delta(c_i, c_j),
\label{eq:modularity}
\end{equation}
where \(A_{ij}\) is an element of the adjacency matrix (1 if an edge exists between nodes \(i\) and \(j\), 0 otherwise), \(k_i\) is the degree of node \(i\), \(m\) is the total number of edges in the graph, \(c_i\) is the community to which node \(i\) is assigned, and \(\delta(c_i, c_j)\) is the Kronecker delta (1 if \(c_i = c_j\), 0 otherwise).

Leveraging these detected communities, HIA refines its node selection by specifically prioritizing:
\begin{itemize}
    \item \textbf{Intra-community High-Attraction Nodes:} High-attraction nodes that exhibit strong temporal degree growth within their own communities.
    \item \textbf{Inter-community Bridge Nodes:} Bridge nodes that demonstrate high betweenness centrality between different communities.
\end{itemize}

This strategic integration of community information ensures that perturbations are context-aware, locally stealthy by respecting community boundaries where appropriate, yet designed for global impact by targeting key inter-community links or influential intra-community hubs. Such a nuanced, community-aware approach enhances the attack's overall effectiveness and potential for evading simple detection mechanisms by considering both micro-level (intra-community) and macro-level (inter-community) graph organization.

The overall perturbation budget, $\Delta$, is determined by a predefined perturbation rate, $\delta$:
\begin{equation}
\label{eq:perturbation_rate_short}
\Delta = \delta \cdot |\mathcal{E}|.
\end{equation}
This budget is then allocated between high-attraction nodes ($\Delta_{\text{high}}$) and bridge nodes ($\Delta_{\text{bridge}}$) using a tunable parameter, $\alpha$:
\begin{equation}
\label{eq:budget_allocation_short}
\Delta_{\text{high}} = \alpha \Delta \quad \text{and} \quad \Delta_{\text{bridge}} = (1 - \alpha) \Delta.
\end{equation}
To prioritize nodes for perturbation, we introduce the $\text{Impact}(v)$ score, which synthesizes the critical factors of a node's importance into a single, actionable metric:

\begin{equation}
\text{Impact}(v) = w_1 \cdot \Delta d_v(t) + w_2 \cdot C_B(v) + w_3 \cdot d_v^{\text{intra}}.
\label{eq:impact_score} 
\end{equation}
Here, the score for node $v$ is a deliberate fusion of its temporal degree growth $\Delta d_v(t)$, which captures its dynamic influence; its betweenness centrality $C_B(v)$, which measures its global structural importance; and its intra-community degree $d_v^{\text{intra}}$, reflecting its local significance. $w_1, w_2, w_3$ are weighting coefficients. We set these to $w_1=0.5, w_2=0.3,$ and $w_3=0.2$, values established via preliminary validation to balance the respective impacts of temporal dynamics ($\Delta d_v(t)$), global centrality ($C_B(v)$), and local community structure ($d_v^{\text{intra}}$). Nodes with higher $\text{Impact}(v)$ scores are prioritized to fill the allocated budgets $\Delta_{\text{high}}$ and $\Delta_{\text{bridge}}$, considering their refined roles (intra-community high-attraction or inter-community bridge).

\subsection{Hybrid Perturbation Implementation}
\label{sec:perturbation}

The final component of HIA involves executing the adversarial modifications through a hybrid approach that combines edge deletion and edge injection \figref{fig:HIA}(C). Crucially, this entire process is guided by the transferable priors learned by the surrogate model. The hybrid strategy leverages the surrogate's edge likelihood predictions to both remove vital connections (high-likelihood edges) and introduce misleading information (low-likelihood edges), thereby maximizing the degradation of the TGNN's predictive performance.

\subsubsection{Edge Deletion}
\label{sec:edge_deletion}

We begin by deleting edges that are deemed essential for maintaining network connectivity and the TGNN's predictive accuracy. The set of deleted edges, $\mathcal{E}_D$, is determined as follows:

\begin{equation}
\label{eq:edge_deletion_short}
\mathcal{E}_D = \{(u, v, t) \in \mathcal{E} \mid \hat{y}_{uvt} > \tau_{\text{del}} \land (\hat{s}_u(t) > \tau_{\text{node}} \lor \hat{s}_v(t) > \tau_{\text{node}})\},
\end{equation}
where $\tau_{\text{del}}$ is a threshold for high edge likelihood. Specifically, it is set as the 85th percentile of the $\hat{y}_{uvt}$ values predicted by the surrogate model for all existing edges in the training dataset, thus targeting edges the surrogate considers highly plausible. Critically, an edge is considered for deletion only if its predicted likelihood is high and it is connected to a high-attraction or bridge node (\secref{sec:node_selection}). This ensures that we are targeting edges that the surrogate model believes are important, and that are connected to nodes that are structurally or dynamically significant. After deletion, the edge set is updated: $\mathcal{E} \leftarrow \mathcal{E} \setminus \mathcal{E}_D$.

\subsubsection{Edge Injection}
\label{sec:edge_injection}

In parallel with edge deletion, we inject new edges to create misleading patterns within the graph. If the edge likelihood score, $\hat{y}_{uvt}$, predicted by the surrogate model is below a threshold, $\tau_{\text{threshold}}$, a new edge, $e_{uv}(t)$, is injected:

\begin{equation}
\label{eq:edge_injection_short}
\hat{y}_{uvt} < \tau_{\text{threshold}} \implies \text{inject edge } e_{uv}(t).
\end{equation}
The threshold $\tau_{\text{threshold}}$ controls the aggressiveness of the injection process. It is set as the 10th percentile of the $\hat{y}_{uvt}$ values predicted by the surrogate model for a large random sample of non-existent node pairs in the training dataset. This data-driven selection aims to identify highly improbable edges for injection, carefully balancing attack effectiveness and stealth. Injecting too many edges could make the attack easily detectable. The set of injected edges is denoted as \(\mathcal{E}_A\). The final perturbed edge set is then constructed as:

\begin{equation}
\label{eq:perturbed_edge_set_short}
\mathcal{E}' = (\mathcal{E} \setminus \mathcal{E}_D) \cup \mathcal{E}_A,
\end{equation}
resulting in the perturbed graph $\mathcal{G}' = (\mathcal{V}, \mathcal{E}', \mathcal{T})$. To maintain stealth and prevent excessive modifications, we enforce a perturbation budget (defined in \eqref{eq:perturbation_rate_short}):

\begin{equation}
\label{eq:perturbation_budget_short}
|\mathcal{E}_A| + |\mathcal{E}_D| \leq \Delta,
\end{equation}
which limits the total number of added and removed edges.

%% file: chapters/Experiments.tex
\section{Experiments}
\label{sec:Experiments}

\begin{table}[t]
\small
\centering
\caption{Statistics of the experimental datasets}
\label{tab:dataset}
\begin{tabular}{lccccc}
\toprule
 & \textbf{WIKI} & \textbf{Reddit} & \textbf{MOOC} & \textbf{LastFM} & \textbf{Bitcoin} \\
\midrule
Nodes & 9K & 11K & 7K & 2K & 5K \\ 
Edges & 157K & 672K & 412K & 1.3M & 35K \\
Time  & $2.7 \times 10^{6}$ & $2.7 \times 10^{6}$ & $2.6 \times 10^{6}$ & $1.3 \times 10^{6}$ & $1.6 \times 10^{8}$ \\
\bottomrule
\end{tabular}
\end{table}

\input{tables/overall-performance}

We evaluate the effectiveness of our proposed HIA on a diverse set of real-world datasets and surrogate models.

\subsection{Experimental Setup}

\subsubsection{Datasets} \label{sec:datasets}

\textbf{WIKI}~\cite{kumar2019predicting} captures interactions on Wikipedia talk pages and contains 9,000 nodes and 157,000 edges over roughly $2.7\times10^6$ timestamped events. \textbf{REDDIT}~\cite{rossi2020temporal} models subreddit hyperlinks with 11,000 nodes, 672,000 edges, and $2.7\times10^6$ time‐ordered interactions. \textbf{MOOC}~\cite{xu2019topology} is a bipartite student–course network featuring 7,000 nodes, 412,000 edges, and $2.6\times10^6$ events. \textbf{LASTFM}~\cite{kumar2019predicting} represents follower relationships among 2,000 users with 1.3 million edges and $1.3\times10^6$ interaction timestamps. \textbf{BITCOIN}~\cite{kumar2019predicting,panzarasa2009patterns} is a trust network from the Bitcoin OTC platform, comprising 5,000 nodes, 35,000 edges, and an extensive $1.6\times10^8$ timestamped transactions.  

For each dataset, we chronologically split edges into 70\% training, 15\% validation, and 15\% testing sets to preserve temporal order.  HIA is applied as a poisoning attack on the training split before model learning.  We remove isolated nodes and duplicate interactions during preprocessing to ensure compatibility with STGNN architectures.

\subsubsection{Evaluated Victim Models} \label{sec:victim_models}

To assess HIA's effectiveness and transferability, we evaluate its performance against four representative TGNN architectures. The selected models encompass diverse mechanisms for temporal processing, embedding aggregation, and memory, allowing for a comprehensive robustness evaluation across different architectural paradigms.
    
\begin{itemize}
    \item \textbf{TGN}~\cite{rossi2020temporal}: A memory-based model using a Gated Recurrent Unit (GRU) to maintain and update node memory, capturing long-range temporal dependencies. HIA disrupts TGN by perturbing key interactions, aiming to corrupt historical node states and degrade predictive performance.
    \item \textbf{JODIE}~\cite{kumar2019predicting}: Another memory-based model employing a Recurrent Neural Network (RNN) to update node embeddings based on past interactions. Its reliance on sequential dependencies makes it vulnerable to targeted disruptions in interaction history.
    \item \textbf{DySAT}~\cite{sankar2020dysat}: A snapshot-based TGNN operating on discrete graph snapshots, applying self-attention both within and across time steps, contrasting with the continuous-time modeling of TGN and JODIE. HIA can disrupt DySAT by perturbing key inter-snapshot transitions, potentially leading to cascading embedding propagation failures.
    \item \textbf{TGAT}~\cite{xu2020inductive}: An attention-based TGNN using attention mechanisms for temporal dependencies without an explicit memory module. HIA targets TGAT by strategically perturbing high-importance edges to mislead attention weights and cause significant information aggregation errors.
\end{itemize}

\subsubsection{Baselines} \label{sec:baselines}

We compare HIA against several baseline attack methods representing a spectrum of adversarial strategies, from naive to more sophisticated, to demonstrate the advantages of our targeted, importance-based approach. These range from simple methods like \textbf{Random}, which deletes or injects edges randomly, to heuristic-based strategies such as \textbf{Preference}~\cite{liben2003link}, which targets high-degree nodes; \textbf{Degree}~\cite{hussain2021structack}, which injects edges between nodes with the fewest common neighbors; and \textbf{PageRank}~\cite{hussain2021structack}, which perturbs nodes with low combined PageRank for stealth. We also compare against state-of-the-art temporal attacks, including \textbf{TGA (Time-aware Gradient Attack)}~\cite{chen2021time}, a white-box, gradient-based method; \textbf{TDAP (Temporal Dynamics-Aware Perturbation)}~\cite{sharma2023temporal}, which maintains temporal consistency; and \textbf{T-Spear}~\cite{lee2024ssaad}, a poisoning attack for continuous-time dynamic graphs.


\subsubsection{Evaluation Protocols} \label{sec:evaluation_protocols}

We evaluate performance on the link prediction task using standard ranking-based metrics, which assess the model's ability to rank true edges higher than non-existent (negative) ones.
\begin{itemize}
    \item \textbf{Mean Reciprocal Rank (MRR)} measures the average inverse rank of the true edge among candidate negative edges. Higher MRR indicates better ranking quality.
    \item \textbf{Hit@10} measures the proportion of cases where the true edge ranks within the top 10 predictions among candidate negative edges.
\end{itemize}
For both metrics, we sample 100 negative edges randomly (without replacement) per positive edge. All experiments are averaged over five random seeds to account for stochasticity in initialization and training.

\subsubsection{Experiment Settings} \label{sec:exp_settings} 
We optimize models using the Adam algorithm with a learning rate of 0.0001, a batch size of 600, and a dropout rate of 0.1.

\subsection{Overall Performance} \label{sec:overall_performance}

Comprehensive results in \tabref{tab:overallresults} consistently demonstrate HIA's superior attack efficacy. Across all five datasets and four TGNN architectures, HIA significantly degrades performance, substantially outperforming all evaluated baselines. For example, against TGN on the WIKI dataset, HIA reduces MRR from a clean 80.5\% to 46.31\%—a far greater degradation than any baseline achieves. Similar significant impacts by HIA are evident across all other dataset-model pairings.

Furthermore, HIA demonstrates strong attack transferability. While victim TGNNs like TGN, JODIE, DySAT, and TGAT utilize diverse temporal encoding mechanisms, they all fundamentally depend on structural connectivity and temporal evolution for link prediction. HIA's consistent success across these architectures indicates it exploits these core, shared dependencies, rather than model-specific idiosyncrasies. By targeting nodes and edges critical to both temporal and structural integrity, HIA disrupts the fundamental information flow relied upon by these varied TGNNs—a broader impact than potentially model-specific baselines. The Average Performance Degradation (A.P.D.) of 35.55\% in MRR across all settings, versus 23.70\% for the best alternative (T-Spear), quantitatively confirms HIA's superior potency and broad applicability against TGNNs.

\subsection{Time Complexity and Scalability of HIA} \label{sec:time_complexity}

HIA achieves scalability via a targeted strategy, focusing computations on a small node subset ($V_{\text{selected}}$) rather than the full graph. After a one-time offline surrogate training, whose computational cost is equivalent to training a standard TGNN on the dataset once, online attack generation involves two main cost components: node selection (approx. $O(|\mathcal{E}| + C_{\text{selection}})$) and the typically dominant perturbation phase. This phase processes edges linked to $V_{\text{selected}}$ (approx. $|V_{\text{selected}}| \cdot d_{\text{selected}}$ edges) in fixed batches ($|B|$), with per-batch costs $O(|B| \cdot C_{\text{tgn}})$ for embeddings and $O(|B|^2 \cdot C_{\text{clf}})$ for likelihoods ($C_{\text{tgn}}$ and $C_{\text{clf}}$ are surrogate/classifier complexities). Since dominant online computations scale with $|V_{\text{selected}}| \ll |\mathcal{V}|$ (total nodes), and $d_{\text{selected}}$ and $|B|$ are bounded, HIA avoids exhaustive full-graph processing per step, enabling efficient scaling to large temporal graphs.

\subsection{Runtime Analysis}
The practical utility of HIA also depends on its computational efficiency. We thus measured average attack generation times (seconds) on WIKI across four TGNN architectures, comparing HIA with baseline strategies (\tabref{tab:runtime_comparison}).

\begin{table}[ht!]
\small
\centering
\caption{Average Attack Generation Time (seconds) on WIKI Dataset across TGNN Architectures.}
\label{tab:runtime_comparison}
\begin{tabular}{@{}lccccc@{}}
\toprule
\textbf{Method} & \textbf{TGN (s)} & \textbf{DySAT (s)} & \textbf{TGAT (s)} & \textbf{JODIE (s)} \\
\midrule\midrule
HIA            & 12      & \textbf{41}        & \textbf{43}       & \textbf{12}        \\
PageRank       & 11      & 65        & 76       & 24        \\
Degree         & 11      & 62        & 72       & 24        \\
Random         & \textbf{10}      & 65        & 73       & 22        \\
Preference     & 23      & 70        & 73       & 25        \\
T-SPEAR        & 22      & 70        & 79       & 25        \\
\bottomrule
\end{tabular}
\end{table}

As shown in \tabref{tab:runtime_comparison}, HIA's runtime is highly efficient. While its generation time on simpler models like TGN and JODIE (12s) is comparable to naive heuristics, it is approximately twice as fast as complex baselines like T-SPEAR. This advantage is magnified on attention-based architectures such as DySAT and TGAT, where HIA is up to 46\% faster than competitors.

%% file: tables/overall-performance.tex
\begin{table*}[h]
\centering
\small
\caption{Comparison of performance (MRR) with HIA across different datasets, surrogate models, and attack methods.}
\label{tab:overallresults}
\begin{tabular*}{\textwidth}{@{\extracolsep{\fill}}ccccccccccc@{}}
\toprule
\textbf{Dataset} & \textbf{Victim Model} & \textbf{Clean} & \textbf{Random} & \textbf{Preference} & \textbf{Degree} & \textbf{PageRank} & \textbf{TGA} & \textbf{TDAP} & \textbf{T-SPEAR} & \textbf{HIA} \\
\midrule\midrule
\multirow{4}{*}{WIKI}
 & TGN   & 80.5 $\pm$ 0.5 & 72.0 $\pm$ 0.6 & 65.3 $\pm$ 1.1 & 66.4 $\pm$ 0.5 & 66.9 $\pm$ 0.9 & 65.2 $\pm$ 0.5 & 63.1 $\pm$ 1.2 & \underline{60.0 $\pm$ 1.6} & \textbf{46.31 $\pm$ 2.1} \\
 & JODIE & 63.2 $\pm$ 1.2 & 46.3 $\pm$ 3.4 & 35.0 $\pm$ 2.3 & \underline{33.0 $\pm$ 2.1} & 35.0 $\pm$ 2.5 & 34.1 $\pm$ 1.5 & 35.4 $\pm$ 0.2 & 33.8 $\pm$ 2.4 & \textbf{21.86 $\pm$ 0.5} \\
 & DySAT & 66.15 $\pm$ 1.2 & 60.18 $\pm$ 3.4 & 54.98 $\pm$ 2.3 & 55.87 $\pm$ 2.1 & 55.94 $\pm$ 2.5 & 52.44 $\pm$ 1.5 & \underline{49.35 $\pm$ 1.3} & 52.31 $\pm$ 2.4 & \textbf{38.47 $\pm$ 0.5} \\
 & TGAT  & 58.58 $\pm$ 0.5 & 48.60 $\pm$ 0.2 & 44.77 $\pm$ 0.9 & 41.17 $\pm$ 0.3 & 40.89 $\pm$ 1.2 & 38.32 $\pm$ 0.5 & 37.44 $\pm$ 1.4 & \underline{35.21 $\pm$ 0.7} & \textbf{23.46 $\pm$ 1.1} \\
\midrule
\multirow{4}{*}{REDDIT}
 & TGN   & 72.61 $\pm$ 1.3 & 51.27 $\pm$ 1.4 & 58.88 $\pm$ 1.2 & 58.52 $\pm$ 1.9 & 55.09 $\pm$ 1.4 & 53.12 $\pm$ 1.5 & 52.43 $\pm$ 1.2 & \underline{51.32 $\pm$ 1.2} & \textbf{49.54 $\pm$ 1.8} \\
 & JODIE & 37.4 $\pm$ 0.0  & 23.8 $\pm$ 0.6 & 29.3 $\pm$ 0.9 & 26.2 $\pm$ 0.8 & 26.6 $\pm$ 0.6 & 26.4 $\pm$ 0.4 & 26.5 $\pm$ 0.1 & \underline{22.8 $\pm$ 0.9} & \textbf{22.53 $\pm$ 1.3} \\
 & DySAT & 42.4 $\pm$ 0.2  & 35.7 $\pm$ 0.3 & 35.2 $\pm$ 0.2 & 35.3 $\pm$ 0.5 & 35.3 $\pm$ 0.3 & 35.4 $\pm$ 0.2 & \underline{33.2 $\pm$ 0.3} & 34.3 $\pm$ 0.1 & \textbf{30.3 $\pm$ 0.1} \\
 & TGAT  & 21.7 $\pm$ 0.1 & 20.2 $\pm$ 0.2 & 20.4 $\pm$ 0.3 & 20.3 $\pm$ 0.4 & 19.5 $\pm$ 0.1 & 19.6 $\pm$ 0.3 & \underline{19.1 $\pm$ 0.2} & 19.2 $\pm$ 0.3 & \textbf{18.15 $\pm$ 0.2} \\
\midrule
\multirow{4}{*}{MOOC}
 & TGN   & 61.8 $\pm$ 1.1 & 55.4 $\pm$ 1.7 & 56.0 $\pm$ 1.6 & 50.3 $\pm$ 1.4 & 53.2 $\pm$ 1.5 & 50.1 $\pm$ 1.3 & 49.2 $\pm$ 1.2 & \underline{48.2 $\pm$ 1.8} & \textbf{38.04 $\pm$ 1.4} \\
 & JODIE & 31.23 $\pm$ 1.2 & 29.03 $\pm$ 1.4 & 29.46 $\pm$ 1.3 & 25.67 $\pm$ 1.1 & 26.93 $\pm$ 1.5 & 25.29 $\pm$ 1.5 & 25.40 $\pm$ 1.4 & \underline{24.3 $\pm$ 1.4} & \textbf{20.29 $\pm$ 1.5} \\
 & DySAT & 18.4 $\pm$ 0.1 & 14.5 $\pm$ 0.1 & 14.5 $\pm$ 0.3 & 14.5 $\pm$ 0.2 & 14.4 $\pm$ 0.2 & 14.3 $\pm$ 0.1 & \underline{14.0 $\pm$ 0.2} & 14.3 $\pm$ 0.2 & \textbf{12.1 $\pm$ 0.2} \\
 & TGAT  & 11.8 $\pm$ 0.1 & 10.2 $\pm$ 0.2 & 10.0 $\pm$ 0.1 & 10.7 $\pm$ 0.2 & 9.6 $\pm$ 0.1 & \underline{9.1 $\pm$ 0.2} & 9.5 $\pm$ 0.2 & 9.3 $\pm$ 0.2 & \textbf{7.2 $\pm$ 0.1} \\
\midrule
\multirow{4}{*}{LASTFM}
 & TGN   & 26.73 $\pm$ 1.3 & 24.24 $\pm$ 1.5 & 23.09 $\pm$ 1.7 & 23.42 $\pm$ 1.8 & 22.88 $\pm$ 1.5 & 23.41 $\pm$ 1.2 & 23.91 $\pm$ 1.4 & \underline{23.09 $\pm$ 1.2} & \textbf{18.58 $\pm$ 1.8} \\
 & JODIE & 9.25 $\pm$ 1.2  & 8.90 $\pm$ 1.3 & 8.52 $\pm$ 1.1 & 8.44 $\pm$ 1.0 & 8.42 $\pm$ 1.3 & 8.10 $\pm$ 1.1 & 8.49 $\pm$ 1.5 & \underline{7.59 $\pm$ 1.5} & \textbf{6.89 $\pm$ 0.5} \\
 & DySAT & 15.39 $\pm$ 0.2 & 12.82 $\pm$ 0.1 & 12.45 $\pm$ 0.5 & 12.37 $\pm$ 0.5 & 12.19 $\pm$ 0.4 & 12.25 $\pm$ 0.2 & 11.59 $\pm$ 0.5 & \underline{11.12 $\pm$ 0.2} & \textbf{10.42 $\pm$ 0.4} \\
 & TGAT  & 9.43 $\pm$ 1.4 & 8.65 $\pm$ 1.3 & 8.24 $\pm$ 1.7 & 8.23 $\pm$ 1.4 & 8.33 $\pm$ 1.3 & 7.48 $\pm$ 1.1 & 7.41 $\pm$ 1.3 & \underline{7.13 $\pm$ 1.8} & \textbf{6.43 $\pm$ 1.5} \\
\midrule
\multirow{4}{*}{BITCOIN}
 & TGN   & 29.31 $\pm$ 0.4 & 25.80 $\pm$ 0.7 & 25.12 $\pm$ 0.3 & \underline{24.12 $\pm$ 1.2} & 24.52 $\pm$ 0.6 & 24.44 $\pm$ 0.9 & 24.28 $\pm$ 0.5 & 24.34 $\pm$ 2.1 & \textbf{20.14 $\pm$ 4.3} \\
 & JODIE & 19.94 $\pm$ 1.2 & 17.65 $\pm$ 0.8 & 16.06 $\pm$ 0.7 & 16.01 $\pm$ 0.8 & 16.05 $\pm$ 0.3 & 16.02 $\pm$ 0.4 & \underline{15.79 $\pm$ 2.5} & 16.04 $\pm$ 0.3 & \textbf{12.90 $\pm$ 0.5} \\
 & DySAT & 63.03 $\pm$ 0.6 & 59.24 $\pm$ 0.7 & 59.11 $\pm$ 0.3 & 59.33 $\pm$ 0.5 & 58.89 $\pm$ 0.5 & 59.80 $\pm$ 0.4 & 58.77 $\pm$ 0.3 & \underline{58.24 $\pm$ 0.2} & \textbf{52.30 $\pm$ 0.3} \\
 & TGAT  & 15.7 $\pm$ 0.3  & 12.2 $\pm$ 0.4 & 11.5 $\pm$ 0.3 & 12.1 $\pm$ 0.3 & 11.1 $\pm$ 0.4 & 11.3 $\pm$ 0.4 & \underline{10.8 $\pm$ 0.3} & 11.5 $\pm$ 0.5 & \textbf{10.3 $\pm$ 0.2} \\
\midrule\midrule
\multicolumn{2}{c}{A.P.D.$\downarrow$} 
& 0\% & -14.69\% & -16.95\% & -18.49\% & -19.13\% & -21.06\% & -22.05\% & -23.70\% & \textbf{-35.55\%} \\
\bottomrule
\end{tabular*}
\begin{minipage}{\textwidth}
\small
\raggedright
1. MRR measured at perturbation rate $p = 0.3$.\\
2. For each victim model, the best and second-best performances are shown in \textbf{bold} and \underline{underlined}, respectively.\\
3. A.P.D.\ denotes average performance degradation.
\end{minipage}
\end{table*}

%% file: chapters/Ablation.tex
\section{Ablation Study}
\label{sec:ablation}


\begin{figure*}[t]
\centering
\includegraphics[width=0.9\textwidth]{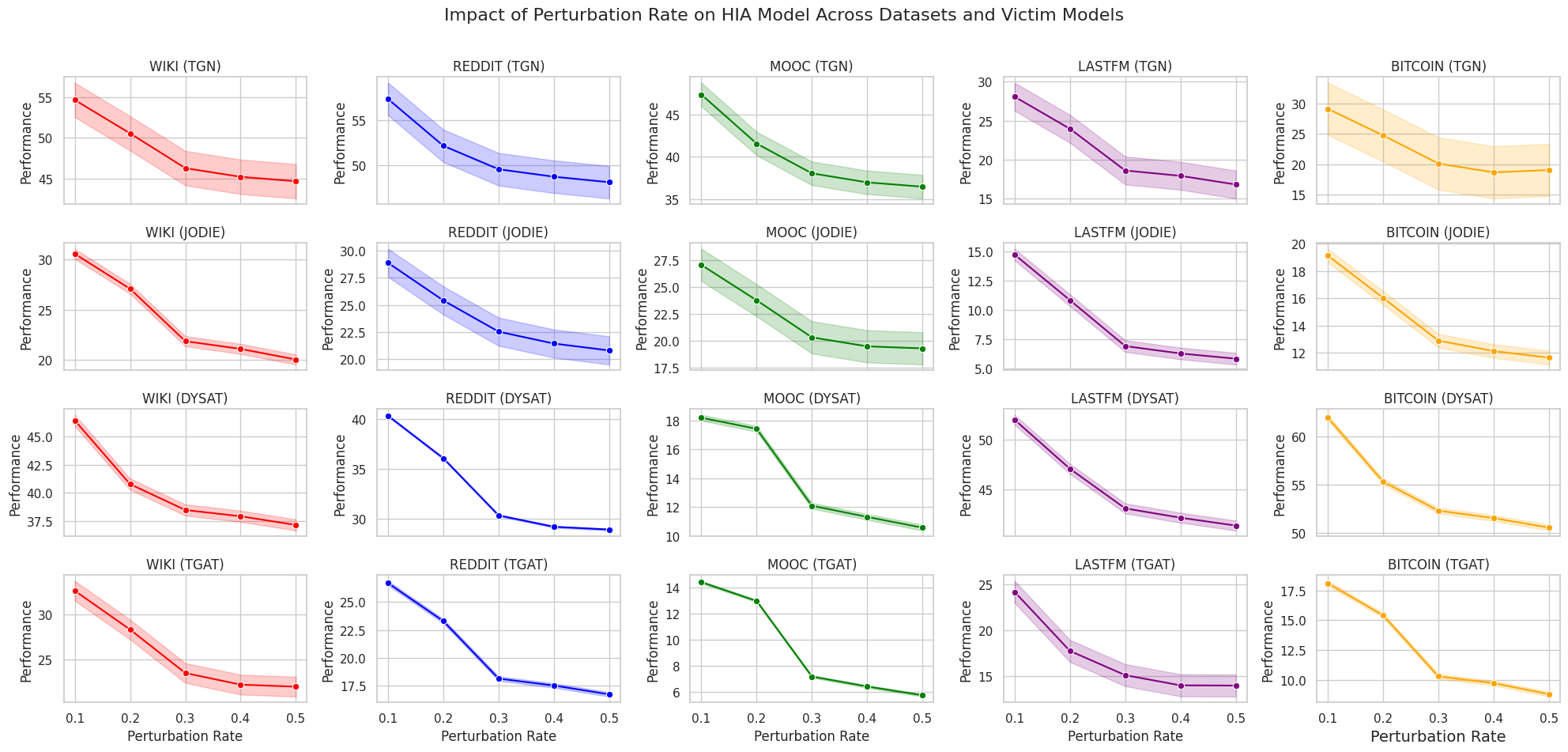}
\caption{MRR vs. Perturbation Rate on HIA. As the perturbation rate increases, both MRR and Hit@10 decrease, demonstrating the attack’s effectiveness. The decline stabilizes beyond 30\%, indicating diminishing returns in performance degradation.}
\label{fig:PerturbationRate}
\end{figure*}
To comprehensively evaluate the contribution of each key component in HIA, we conduct a series of ablation studies. Unless otherwise specified, these studies use the WIKI dataset, the TGN surrogate model, MRR as the primary performance metric, and a default perturbation rate of $\delta = 0.3$. Key WIKI-specific results are consolidated in \tabref{tab:combined_ablation}, with detailed discussions for each study below.

\subsection{Analysis on Perturbation Rate and Attack Stealth}
\label{sec:ablation_perturbation_rate}
The perturbation rate $\delta$ is a critical parameter governing the trade-off between attack efficacy and detectability. As shown in \figref{fig:PerturbationRate}, HIA’s impact grows with an increasing budget, with performance degradation saturating near $\delta=0.3$. This indicates a point of diminishing returns, making it an efficient default rate. HIA's strategic nature ensures substantial impact even at low budgets; at just 10\% perturbation on WIKI, it significantly outperforms naive baselines.

Crucially, HIA achieves this high impact while maintaining stealth. We assessed this on WIKI using two metrics shown in \tabref{tab:stealth}: structural deviation ($\mathrm{KL}_{d}$) and temporal consistency (\textit{TimeCross}). The results confirm HIA’s unobtrusiveness; its structural deviation ($\mathrm{KL}_{d}=0.014$) is minimal and comparable to a random attack, while its temporal violations ($\mathrm{TimeCross}=0.4\%$) are significantly lower than other potent attacks like T-SPEAR. This demonstrates that HIA’s importance-guided strategy successfully executes powerful yet plausible modifications, balancing high efficacy with low detectability.




\begin{table}[h]
\small
\centering
\caption{Stealth metrics on WIKI @ $\delta=0.3$ (lower values indicate greater stealth).}
\label{tab:stealth} 
\begin{tabular}{@{}lcccc@{}} 
\toprule
\textbf{Metric} & \textbf{Random} & \textbf{Degree} & \textbf{T-SPEAR} & \textbf{HIA} \\
\midrule\midrule
$\mathrm{KL}_{\!d}$ & 0.012 & 0.019 & 0.025 & 0.014 \\
$\mathrm{TimeCross}$ (\%) & 0.0 & 0.0 & 1.3 & 0.4 \\
\bottomrule
\end{tabular}
\end{table}


\begin{table}[htb]
\small
\centering
\caption{Combined ablation results on WIKI (TGN surrogate)}
\label{tab:combined_ablation}
\begin{tabular}{@{}p{0.49\columnwidth} p{0.34\columnwidth} r@{}}
\toprule
\textbf{Category} & \textbf{Variation} & \textbf{MRR} \\
\midrule\midrule
\multirow{2}{*}{(a) Community Detection (CD)}
 & With CD (HIA default) & 46.31 \\
 & Without CD            & 50.12 \\
\midrule
\multirow{3}{*}{(b) Hybrid Perturbation}
 & Hybrid (HIA default)  & 46.31 \\
 & Injection only        & 52.45 \\
 & Deletion only         & 50.88 \\
\midrule
\multirow{3}{*}{(c) Node Selection Strategy}
 & HIA (Impact score)    & 46.31 \\
 & Random selection      & 55.93 \\
 & Degree selection      & 63.39 \\
\bottomrule
\end{tabular}
\end{table}

\subsection{Hyperparameter Justification and Robustness}
\label{sec:ablation_hyperparameters}

\begin{table}[h!]
\small
\centering
\caption{Performance of simple node-selection heuristics (Random, Degree) across datasets (MRR)}
\label{tab:node_selection_all_datasets}
\begin{tabular}{l@{\hspace{1cm}}c@{\hspace{1cm}}c@{\hspace{1cm}}c}
\toprule
\textbf{Dataset} & \textbf{Random} & \textbf{Degree} & \textbf{HIA} \\
\midrule\midrule
WIKI    & 55.93 & 63.39 & 46.31 \\
REDDIT  & 68.31 & 71.76 & 49.54 \\
MOOC    & 46.71 & 42.18 & 38.04 \\
LASTFM  & 23.02 & 16.21 & 18.58 \\
BITCOIN & 26.91 & 26.62 & 20.14 \\
\bottomrule
\end{tabular}
\end{table}

While $\delta$ is the primary budget parameter, other hyperparameters like the temporal window $\Delta t$ were set based on preliminary validation and established practices to ensure robust performance. The generalizability of these choices is confirmed by HIA's consistent state-of-the-art performance across all datasets (\tabref{tab:node_selection_all_datasets}) and the stable contributions of its core components in ablation studies (\tabref{tab:combined_ablation}).

The results in \tabref{tab:node_selection_all_datasets} reveal \emph{why} HIA’s multi-faceted strategy is superior to simpler heuristics: the effectiveness of naive \texttt{Degree} or \texttt{Random} attacks is highly dependent on specific graph topologies. For instance, \texttt{Degree} attacks are ineffective on WIKI and REDDIT because they target saturated legacy hubs, but they are more potent on the sparse, seasonal structure of MOOC. Similarly, \texttt{Random} attacks fail on sparse graphs like BITCOIN where perturbations must be precise. HIA’s $\text{Impact}(v)$ score, in contrast, consistently outperforms these rigid heuristics by adapting to each graph's unique properties. It deliberately fuses \textbf{temporal growth} to catch WIKI's emerging influencers, \textbf{betweenness centrality} to sever BITCOIN's critical trust bridges, and \textbf{community context} to disrupt LASTFM's clusters, making it a robust and universally effective attack strategy.



\subsection{Optimizing Targeting: Synergistic Disruption of Attraction and Bridge Nodes}
\label{sec:ablation_attraction_bridge}

HIA strategically allocates its perturbation budget $\Delta$ using parameter $\alpha$ between high-attraction nodes (critical for local temporal influence) and bridge nodes (essential for global structural connectivity). We analyzed HIA's synergistic approach against strategies targeting only one node type.

On the WIKI dataset, targeting only high-attraction nodes ($\alpha=1$) yielded an MRR/Hit@10 of 46.83/71.96, while targeting only bridge nodes ($\alpha=0$) resulted in 47.90/71.75. HIA's empirically optimized default configuration, however, sets $\alpha=0.6$ (allocating 60\% of the budget to high-attraction nodes, $\Delta_{\text{high}}$, and 40\% to bridge nodes, $\Delta_{\text{bridge}}$), achieving a superior MRR/Hit@10 of 46.31/70.40.

This balanced strategy improved MRR by 1.1--3.32\% and Hit@10 by 1.35--1.56 points over these single-target approaches. HIA's strength lies in its calibrated, dual-pronged disruption of complementary pathways—local dynamic influence and global structural connectivity. This concurrent targeting prevents TGNNs from compensating for singular vulnerabilities, making the nuanced allocation critical to HIA's enhanced effectiveness over simpler schemes.

\subsection{Impact of Community Detection and Hybrid Perturbation}
\label{sec:ablation_community}
The value of community-aware targeting is validated by comparing HIA with a variant lacking community detection (CD). As \tabref{tab:combined_ablation}(a) shows, omitting CD impairs attack efficacy (MRR 50.12 vs. HIA's 46.31). This indicates that tailoring perturbations by considering community structures—such as prioritizing inter-community bridges or influential intra-community nodes as per \eqref{eq:impact_score}—enhances disruptive potential by exploiting the graph's meso-scale organization.

Further, the hybrid edge injection/deletion strategy is critical. \tabref{tab:combined_ablation}(b) demonstrates that HIA's hybrid approach (MRR 46.31) significantly outperforms relying solely on edge injection (MRR 52.45) or edge deletion (MRR 50.88). This highlights the synergistic benefit of concurrently removing vital connections and introducing specious ones to maximize TGNN performance degradation.

%% file: chapters/Conclusion.tex
\section{Conclusion and Future Work} \label{sec:Conclusion}

We have presented HIA, a restricted black-box attack framework that systematically degrades the link-prediction performance of STGNNs. By ranking node–time pairs via a data-driven surrogate and applying our hybrid of edge injections and deletions, HIA concentrates perturbations on those elements most critical to both graph structure and temporal evolution. Empirically, HIA achieves up to a 35.55\% drop in MRR across multiple datasets and architectures, revealing a previously underexplored vulnerability in STGNNs.

These findings underscore the urgent need for more robust TGNNs and highlight several key directions for future research. Critical next steps include evaluating HIA against existing defense mechanisms and adapting its principles for real-time attack generation in streaming scenarios. The attack strategy could be enhanced by dynamically allocating the perturbation budget based on evolving graph properties and investigating different surrogate model architectures for improved transferability.

Beyond extending the attack, our results illuminate paths toward building more robust defenses. Adversarial training regimes must incorporate multi-step attack sequences to withstand cascading perturbations. Dynamic graph purification techniques should distinguish genuine temporal changes from adversarial edits by learning normal evolution patterns. Finally, integrating node-importance estimates into training can mitigate attacks targeting critical elements. These directions will help close the vulnerability gap exposed by HIA, moving toward STGNNs that are robust against spatio-temporal adversaries.

\begin{acks}
This study is partially supported by NIH grant R01AG084236, \newline R01AG083039. This work was supported in part by the U.S. National Science Foundation under Grant No. 2317117.  
\end{acks}

%% file: chapters/GenAIDisclosure.tex
\section*{GenAI Usage Disclosure}

The authors acknowledge the use of generative AI tools in the preparation of this manuscript. Specifically, Google's Gemini large language model was utilized during various stages of the research writing process. 

The assistance provided by the GenAI tool included, but was not limited to:
\begin{itemize}
    \item Assisting in the editorial refinement of the manuscript, including improving clarity, conciseness, academic tone, grammar, and punctuation of author-drafted text, as well as rephrasing sentences and paragraphs to strengthen arguments based on the authors' core ideas and provided information.
    \item Generating LaTeX code for textual revisions and table structures based on content provided by the authors.
\end{itemize}

The authors directed all analyses and interpretations presented. While the GenAI tool assisted in drafting and refining the language and structure of the text, the core ideas, experimental design, conduct, results, and final manuscript content were conceived, executed, and validated by the human authors. All AI-generated text or suggestions were critically reviewed, edited, and revised by the authors to ensure accuracy, consistency with the authors' research, and to reflect their unique intellectual contributions. The authors take full responsibility for all content in this work, including any parts where GenAI assistance was used in the drafting process, in accordance with ACM's Authorship Policy.